\documentclass[preprint,12pt]{elsarticle}

\usepackage{url} 
\usepackage{amsmath} 
\usepackage{flushend}
\usepackage{booktabs} 
\usepackage{algorithm} 
\usepackage{algpseudocode}
\usepackage{graphicx}
\usepackage{fullpage}
\usepackage{caption}
\usepackage{xcolor}
\usepackage{subfigure}
\usepackage{longtable}
\usepackage{tablefootnote}
\usepackage{hhline}
\usepackage{footnote}
\usepackage{supertabular}
\usepackage{array}
\usepackage[flushleft]{threeparttable}
\usepackage{multirow}

\usepackage{lscape}
\usepackage{enumitem}
\usepackage{textcomp} 
\usepackage{pifont}
\usepackage{adjustbox}
\usepackage{makecell}
\usepackage{verbdef}
\usepackage{colortbl}
\usepackage{ltablex,booktabs}
\usepackage{xspace}
\usepackage[normalem]{ulem}
\usepackage{lineno}
\usepackage{tikz}
\usepackage{color}
\biboptions{authoryear}

\journal{Preprint. Under review.}

\begin{document}

\begin{frontmatter}
\title{Prediction of Football Player Value using Bayesian Ensemble Approach}

\author[label1]{Hansoo Lee}
\author[label2]{Bayu Adhi Tama$^{\star}$}
\author[label1]{Meeyoung Cha$^{\star}$}

\address[label1]{School of Computing, KAIST, Daejeon, Republic of Korea}
\address[label2]{Department of Information Systems, University of Maryland, Baltimore County, MD, USA\\
$^{\star}$Corresponding author}

\begin{abstract}
The transfer fees of sports players have become astronomical. This is because bringing players of great future value to the club is essential for their survival. We present a case study on the key factors affecting the world's top soccer players' transfer fees based on the FIFA data analysis. To predict each player's market value, we propose an improved LightGBM model by optimizing its hyperparameter using a Tree-structured Parzen Estimator (TPE) algorithm. We identify prominent features by the SHapley Additive exPlanations (SHAP) algorithm. The proposed method has been compared against the baseline regression models (e.g., linear regression, lasso, elastic net, kernel ridge regression) and gradient boosting model without hyperparameter optimization. The optimized LightGBM model showed an excellent accuracy of approximately 3.8, 1.4, and 1.8 times on average compared to the regression baseline models, GBDT, and LightGBM model in terms of RMSE. Our model offers interpretability in deciding what attributes football clubs should consider in recruiting players in the future.
\end{abstract}

\begin{keyword}
Soccer\sep Football\sep Sports analytics\sep Correlation analysis\sep Ensemble learning\sep Hyperparameter optimization\sep Gradient boosting model\sep Lightgbm model
\end{keyword}

\end{frontmatter}

\section{Introduction}

Prediction of future market players' value is important because it could seriously burden professional football clubs. On the other hand, it allows clubs to gain profit by selling a well-performing player at a high price. Famous clubs such as FC Barcelona and Manchester United spend astronomical values to obtain the best players. Hence, identifying the factors affecting football players' value might bring competitive advantages to small and medium-sized football clubs.    

Previous studies indicated that the player's market value factors have varied, such as demographic information, profile, and real performance through the sports statistics sites (e.g., WhoScored\footnote{https://1xbet.whoscored.com/})~\cite{he2015football, muller2017beyond, cwiklinski2021will}. However, the sports statistics site has no detailed information about players' quantitative ability information (e.g., attack, goalkeeping, defense, mental). Hence, there is a limit to grasping how football ability factors affect the player's market value using the sports statistic sites. Meanwhile, football video game data from EA Sports\footnote{https://www.ea.com/sports}'s SOFIFA\footnote{https://sofifa.com/} and Football Manager\footnote{https://www.footballmanager.com/} can be used as an alternative to overcome the limitations of the existing studies~\cite{yiugit2019football, yigit2020xgboost}.
Among the video game data, the SOFIFA dataset comprises approximately 55 attributes related to each player's ability (e.g., passing, attacking), position (e.g., goal keeper, midfielder), demographic information (e.g., age, height, weight), monetary value (e.g., wage, release clause), and profile (e.g., club name, international reputation). To make a reliably quantified dataset of all players, EA Sports employs 30 EA producers and 400 outside data contributors who are responsible for ensuring all player data is up to date, while a community of over 6,000 SOFIFA data reviewers or talent scouts from all over the world is constantly providing suggestions and alterations to the database. Next, EA Sports employees build FIFA's game ability attributes dataset every year based on a fair evaluation of more than 30 leagues, more than 700 clubs, and more than 17,000 players and update the dataset every month based on the actual competition performance of the competitors. For example, EA Sports’ staff watch every game to find out the pace, not even the major league, but the players in second division leagues such as English Football League (EFL) Championship. 

Therefore, although SOFIFA dataset is video game data, SOFIFA dataset provides all objectively evaluated ability attributes of all players, and quantified ability data of more than 17,000 players, which is the most reliable among all football-related data via reflecting real player stats. Because of the above advantages, SOFIFA dataset is being used for various research purposes (e.g., match results, player's market value prediction, clustering player's position, and player's performance prediction)~\cite{rajesh2020data, behravan2020novel, pariath2018player, prasetio2016predicting, soto2017gaussian} as shown in Table~\ref{sofifapaper}. However, SOFIFA dataset does not provide the club's match record attributes such as win/draw/lose rates and goal/assist points, which are considered by previous studies~\cite{cwiklinski2021will, he2015football, muller2017beyond} from WhoScored dataset as shown in Table~\ref{existingpaper} Furthermore, while not covered in previous studies, WhoScored also provides player match attributes such as violation records (e.g., foul, card), number of game played, and attack point records (e.g., goal, assist, affective shooting). 


%

In terms of market value prediction model, Existing studies mainly rely on weak regression techniques such as linear regression, regularized regression (e.g., ridge, lasso), and regression tree for player's market value prediction~\cite{he2015football, muller2017beyond} as shown in Table~\ref{existingpaper}. However, ~\cite{behravan2020novel} used clustering technique, while other well-known ensemble learners, e.g., AdaBoost and XGBoost were also applied to improve the model’s performance in this domain prediction~\cite{yigit2020xgboost, cwiklinski2021will, yiugit2019football} as shown in Table~\ref{existingpaper}, \ref{sofifapaper}


Through this review, we find three research gaps in the existing studies. First, no previous studies simultaneously have used the both of SOFIFA and WhoScored attributes to predict market value (i.e., existing studies used only either SOFIFA or WhoScored dataset). Second, the importance of features of each predictive model is not defined in the existing studies, making it difficult to determine which features can predict the player's market values well. Third, no studies considered hyperparameter optimization to improve the model's performance for predicting players’ market value. Fourth, most existing football data-driven studies are not considered about decreasing learning time and enhancement of performance by using state-of-the-art model and hyperparameter optimization technologies (e.g., lightGBM, TPE bayesian optimization).


In this paper, we first extract the attributes from both SOFIFA and WhoScored dataset which are the player's all ability, profile, demographic information, and monetary value attributes provided by SOFIFA and club \& individual player's match record attributes in belong to big five major European soccer leagues (English Premier League, Spain La Liga, Germany Bundesliga, France Ligue 1, Italy Serie A) from WhoScored. Then, the state-of-the-art ensemble model (i.e., optimized LightGBM model via Bayesian optimization) is utilized to analyze the causal relationship between factors that contribute to future player's market value prediction. The prediction accuracy of the model is validated in terms of root mean square error (RMSE) and mean absolute error (MAE) metrics with cross-validation. Finally, we identified the features importance by SHAP value, and derived the best features which can explain the market value prediction model best among all attributes.

To sum up, the main objective of this paper are as follows.   
\begin{itemize}
    \item We crawled and extracted all types of attributes related to player's ability or club's match record from two data sources (e.g., SOFIFA, WhoScored) to identify all possible features affecting market value prediction that were not considered in previous studies.
    \item We develop a predictive optimized ensemble model (e.g., LightGBM + TPE Bayesian optimization) that can predict player's market value accurately.
    \item We seek the features that have significant impacts on predicting players' market value.
\end{itemize}




\begin{table*}
\centering
\caption{Previous research with market value prediction using real data source}
\label{existingpaper}
\resizebox{1\textwidth}{!}{%
\begin{savenotes}
\begin{tabular}{@{}lllll@{}}

\toprule
\multicolumn{1}{c}{\textbf{Ref.}} &
  \multicolumn{1}{c}{\textbf{Research Purpose}} &
  \multicolumn{1}{c}{\textbf{Data Source}} &
  \multicolumn{1}{c}{\textbf{Features}} &
  \multicolumn{1}{c}{\textbf{\begin{tabular}[c]{@{}c@{}}Modeling \\ Technique\end{tabular}}} \\ \midrule\midrule
{\textbf{\cite{he2015football}}} &
  \textbf{\begin{tabular}[c]{@{}l@{}}Estimation of player's \\ performance and market value,\\ and relationship between \\ player's performance and \\ market value by regression model\end{tabular}} &
  \textbf{{\begin{tabular}[c]{@{}l@{}}Transfer Market\footnote{https://www.transfermarkt.co.uk/wettbewerbe/national},\\ WhoScored, \\ European Football \\ Database\footnote{https://www.footballdatabase.eu/en/}, \\ and Garter\footnote{https://www.theguardian.com/football/2014/dec/21/how-the-guardian-rankedthe-2014-worlds-top-100-footballers}\end{tabular}}} &
  \textbf{{\begin{tabular}[c]{@{}l@{}}transfer fee, \\performance assessments, \\age, contract duration\end{tabular}}} &
  \textbf{Lasso Regression} \\
  \midrule
{\textbf{\cite{majewski2016identification}}} &
  \textbf{\begin{tabular}[c]{@{}l@{}}Estimation of player's \\ market value and identifying \\ the determining factors of\\ market value by regression model\end{tabular}} &
  {\textbf{Transfer Market}} &
  \textbf{\begin{tabular}[c]{@{}l@{}} 5 Human capital factors\\ (e.g., age),\\ 5 Productivity factors\\(e.g., goals scored),\\ 4 Organizational capital factors\\(e.g., total time) \end{tabular}} &
  \textbf{Linear Regression} \\
  \midrule
{\textbf{\cite{muller2017beyond}}} &
  \textbf{\begin{tabular}[c]{@{}l@{}}Estimation of player's \\ market value by regression model\end{tabular}} &
  \textbf{{\begin{tabular}[c]{@{}l@{}}Google\footnote{https://www.google.com/},\\ Reddit\footnote{https://www.reddit.com/}, \\ Transfer Market,\\ WhoScored, \\ Wikipedia\footnote{https://en.wikipedia.org/wiki/Main\_Page},\\ Youtube\footnote{https://www.youtube.com/}\end{tabular}}} &
  \textbf{\textbf{\begin{tabular}[c]{@{}l@{}} 1 Player valuation\\(e.g., market value), \\ 3 Player characteristics \\(e.g., Age),\\16 Player Performance\\(e.g., Minutes played),\\ 4 Player popularity\\(e.g., Wikipedia page views)\end{tabular}}} &
  \textbf{Linear Regression} \\
  \midrule
{\textbf{\cite{cwiklinski2021will}}} &
  \textbf{\begin{tabular}[c]{@{}l@{}}Supporting a football team building \\ and successful player's transfer \\ by classification model\end{tabular}} &
  \textbf{{\begin{tabular}[c]{@{}l@{}}WhoScored,\\ TransferMarket, \\ Sofascore~\footnote{https://www.sofascore.com/}\end{tabular}}} &
  \textbf{\textbf{\begin{tabular}[c]{@{}l@{}} 4 Physical parameters \\(e.g., matches played),\\ 28 Technical parameters \\(e.g., Goals from the penalty box),\\ 6 Psychological parameters \\ (e.g, Age)\end{tabular}}} &
  \textbf{\begin{tabular}[c]{@{}l@{}}Random Forest, Naive Bayes, \\ and AdaBoost\end{tabular}} \\ \bottomrule
\end{tabular}%
\end{savenotes}
}
\end{table*}

\begin{table*}
\centering
\caption{Previous research with diverse research purpose using game data source along with real data source}
\label{sofifapaper}
\resizebox{\textwidth}{!}{%
\begin{savenotes}
\begin{tabular}{@{}lllll@{}}

\toprule
\multicolumn{1}{c}{\textbf{Ref.}} &
  \multicolumn{1}{c}{\textbf{Research Purpose}} &
  \multicolumn{1}{c}{\textbf{Data Source}} &
  \multicolumn{1}{c}{\textbf{Features}} &
  \multicolumn{1}{c}{\textbf{\begin{tabular}[c]{@{}c@{}}Modeling \\ Technique\end{tabular}}} \\ \midrule\midrule
{\textbf{\cite{prasetio2016predicting}}} &
  \textbf{\begin{tabular}[c]{@{}l@{}}Estimation of match results\\ by classification model\end{tabular}} &
  \textbf{\begin{tabular}[c]{@{}l@{}}Match records of \\ Premier League\footnote{http://www.football-data.co.uk/} \\ and SOFIFA\end{tabular}} &
  \textbf{\begin{tabular}[c]{@{}l@{}}Match records season \\ 2010/2011-2015/2016 \\ in terms of 4 variables: \\ Home Offense, Home Defense, \\ Away Offense, and Away Defense\end{tabular}} &
  \textbf{\begin{tabular}[c]{@{}l@{}}Logistic Regression model \\ using Newton-Raphson algorithm\end{tabular}} \\
  \midrule
{\textbf{\cite{pariath2018player}}} &
  \textbf{\begin{tabular}[c]{@{}l@{}}Estimation of player's overall \\ performance and market value \\ by regression model\end{tabular}} &
  \textbf{{SOFIFA}} &
  \textbf{\begin{tabular}[c]{@{}l@{}}Approximately 36 attributes:\\ Physical (e.g., Age),\\  Attacking (e.g., finishing),\\ Movement (e.g., acceleration),\\ Skill (e.g., dribbling),\\ Defensive (e.g., Marking),\\ Mentality (e.g., aggression),\\ Power (e.g., jumping),\\ General (e.g., overall rating)\end{tabular}} &
  \textbf{Linear Regression} \\
  \midrule
{\textbf{\cite{rajesh2020data}}} &
  \textbf{\begin{tabular}[c]{@{}l@{}}Estimation of player’s position \\ by classification model\\ and Clustering player's positions \\ by age and overall performance \\ by clustering model\end{tabular}} &
  \textbf{{SOFIFA}} &
  \textbf{\begin{tabular}[c]{@{}l@{}}Approximately 35 attributes:\\1 Physical (e.g., BMI), \\5 Attacking (e.g., finishing), \\Movement (e.g., acceleration),\\ 5 Skill (e.g., dribbling), \\3 Defensive (e.g., Marking),\\ 5 Mentality (e.g., aggression),\\ 2 Monetary value (e.g., wage),\\ 4 General (e.g., potential),\\ 5 Power (e.g., shot power)\end{tabular}} &
  \textbf{\begin{tabular}[c]{@{}l@{}}Naïve Bayes, Decision Tree, \\ Random forest, SVC\end{tabular}} \\
  \midrule
{\textbf{\cite{soto2017gaussian}}} &
  \textbf{\begin{tabular}[c]{@{}l@{}}Clustering football players' \\position by clustering model\end{tabular}} &
  \textbf{{SOFIFA}} &
  \textbf{\begin{tabular}[c]{@{}l@{}}Approximately 40 attributes:\\ 4 Physical (e.g., Weight),\\ 5 Attacking (e.g., Crossing),\\ 5 Movement (e.g., Agility),\\ 5 Skill (e.g., Curve),\\ 3 Defensive (e.g., Tackle),\\ 5 Mentality (e.g., Positioning),\\ 5 Goalkeeping (e.g., Diving),\\ 5 Power (e.g., jumping),\\ 2 General (e.g., potential)\end{tabular}} &
  \textbf{\begin{tabular}[c]{@{}l@{}}Gaussian mixture \\model-based clustering, \\ XGBoost for classification\end{tabular}} \\
  \midrule
{\textbf{\begin{tabular}[c]{@{}l@{}}\cite{behravan2020novel}\end{tabular}}} &
  \textbf{\begin{tabular}[c]{@{}l@{}}Estimation of player's market\\ value by regression model\end{tabular}} &
  \textbf{{SOFIFA}} &
  \textbf{\begin{tabular}[c]{@{}l@{}}In 55 attributes \\ (Physical, Attacking,\\ Movement, Skill, Defensive,\\ Mentality, Power, General),\\ 5, 32, 30, and 28 features\\ were selected for goalkeeper,\\ strikers, defenders,\\ and midfielders positions \\by PSO clustering,\\ respectively.\end{tabular}} &
  \textbf{\begin{tabular}[c]{@{}l@{}} Particle Swarm Optimization\\(PSO) SVR,\\Gery Wolf Optimizer\\(GWO) SVR,\\Inclined Planes\\ System Optimization\\(IPO) SVR,\\Whale Optimization Algorithm\\(WOA) SVR\end{tabular}} \\
  \midrule
{\textbf{\cite{yiugit2019football, yigit2020xgboost}}} &
  \textbf{\begin{tabular}[c]{@{}l@{}}Estimation of player's \\ market value by regression model\end{tabular}} &
  \textbf{{\begin{tabular}[c]{@{}l@{}}Football Manager, \\ Transfer Market\end{tabular}}} &
  \textbf{\begin{tabular}[c]{@{}l@{}}4 main chapters which are; \\technical, mental, physical, \\and goalkeeping \\with 49 attributes\end{tabular}} &
  \textbf{\begin{tabular}[c]{@{}l@{}}linear regression,\\ ridge regression, \\ lasso regression, \\principal component regression,\\ random forest, XGBoost\end{tabular}} \\
 \bottomrule
\end{tabular}%
\end{savenotes}
}
\end{table*}

\section{Proposed Method}
\label{propmeth}
In this section, baseline models and the proposed model is briefly described.
\subsection{Regularized Linear Regression Model}
As the baseline model, we use various linear regression models. 
First, we used multiple linear regression model; a representative linear regression model (LM). It is commonly used when a dependent variable and more than two independent variables exist. To avoid overfitting due to variances in the dataset, we also use other baseline regression models, i.e., lasso~\cite{tibshirani1996regression}, kernel trick-based ridge regression (KRR)~\cite{welling2013kernel}, and elastic net (E-Net) regularization~\cite{zou2005regularization}.

\subsection{Gradient Boosting Decision Tree Model}
We used a gradient boosting decision tree (GBDT)~\cite{friedman2001greedy} as another baseline model. As one of boosting algorithms, GBDT, is generally known to show the outperformed accuracy compared to other machine learning algorithms and bagging ensemble learning models such as random forest~\cite{hastie2009elements}. Therefore, we use the GBDT methods for the baseline ensemble learning model via the scikit-learn API.

\subsection{LightGBM}
LightGBM is an improved variant of the most state-of-the-art GBDT algorithm introduced in Ke et al. ~\cite{ke2017lightgbm}. LightGBM generally possesses high efficiency (i.e., fast training while maintaining high performance) compared to GBDT and XGBoost on high dimensional data ~\cite{ke2017lightgbm}. Another advantage of LightGBM is that unlike other boosting algorithms that require numerical transformation (e.g., label encoding, one-hot encoding), it handles categorical features internally using the grouping method~\cite{fisher1958grouping}. 

Therefore, when using LightGBM, the data pre-processing can be shortened because numerical transformation and normalization of features are unnecessary. LightGBM adopts a leaf-wise tree generation strategy that can reduce losses more than the traditional level-wise strategy when the leaf grows. Therefore, the final model of LightGBM is composed of a smaller number of decision trees and a smaller number of leaves per decision tree, enabling efficient matching and time in the decision-making process. Based on these advantages, in this study, we employ LightGBM to increase the learning speed while maintaining excellent performance for football player's market value.

\subsection{Hyperparameter Optimization}
We used Bayesian optimization with a tree-structured Parzen estimator approach (TPE) as a hyperparameter optimization (HPO) algorithm. Unlike other black-box optimization (BBO) methods (e.g., grid and random search), Bayesian optimization forms a probabilistic model that maps hyperparameters to the objective's score probabilities function~\cite{bergstra2011algorithms}. Therefore, Bayesian optimization can find better hyperparameters in less contrast time compared with other BBOs. 

The formalization of Bayesian optimization is sequential model-based optimization (SMBO). Surrogate models (i.e., the model that determines with which evaluation points are fitted) affect SMBO results, including Gaussian processes, random forest regression, and TPE. TPE is known to be more flexible than traditional Bayesian optimization \cite{bergstra2011algorithms}. Besides, when the TPE algorithm is used in HPO, it shows better accuracy than manual search, Bayesian optimization with Gaussian processes, particle swarm optimization, Nelder-mead procedure, and random search~\cite{olof2018comparative, bergstra2011algorithms}. For the above reasons, we adopt Bayesian optimization with the TPE algorithm for HPO. In this study, we use a state-of-the-art HPO framework called Optuna~\cite{akiba2019optuna}. It is found to be better than Hyperopt~\cite{bergstra2013making} w.r.t ease of use, search space, callback, run pruning, and visualization. In Optuna, we experiment with various conditions, including two TPE algorithms (i.e., independent TPE and multivariate TPE), the Optuna's pruning function (i.e., pruning function can reduce the HPO time with maintaining the performance for the LightGBM model) and also compare with not-used condition.



\section{Experiment}
\label{exp}

\subsection{Data Preprocessing and Feature Extraction}
Dataset is consist of the 2022 SOFIFA dataset provided by sofifa.com and the 2021–2022 ranking table of big five major European soccer leagues by WhoScored. First of all, we performed the data pre-processing in data selection, noise handling, data merging, data grouping, and data transformation. 2,720 players who belong to the 20 top division European clubs (i.e., a club ranked from 1st to 4th in each five European leagues) of 2021–2022 Union of European Football Association (UEFA) Champions League\footnote{https://www.uefa.com/uefachampionsleague/} were selected from the whole player list of 2022 SOFIFA dataset and get rid of missing values in all columns.

\begin{figure*}
  \centering
  \includegraphics[width=\textwidth]{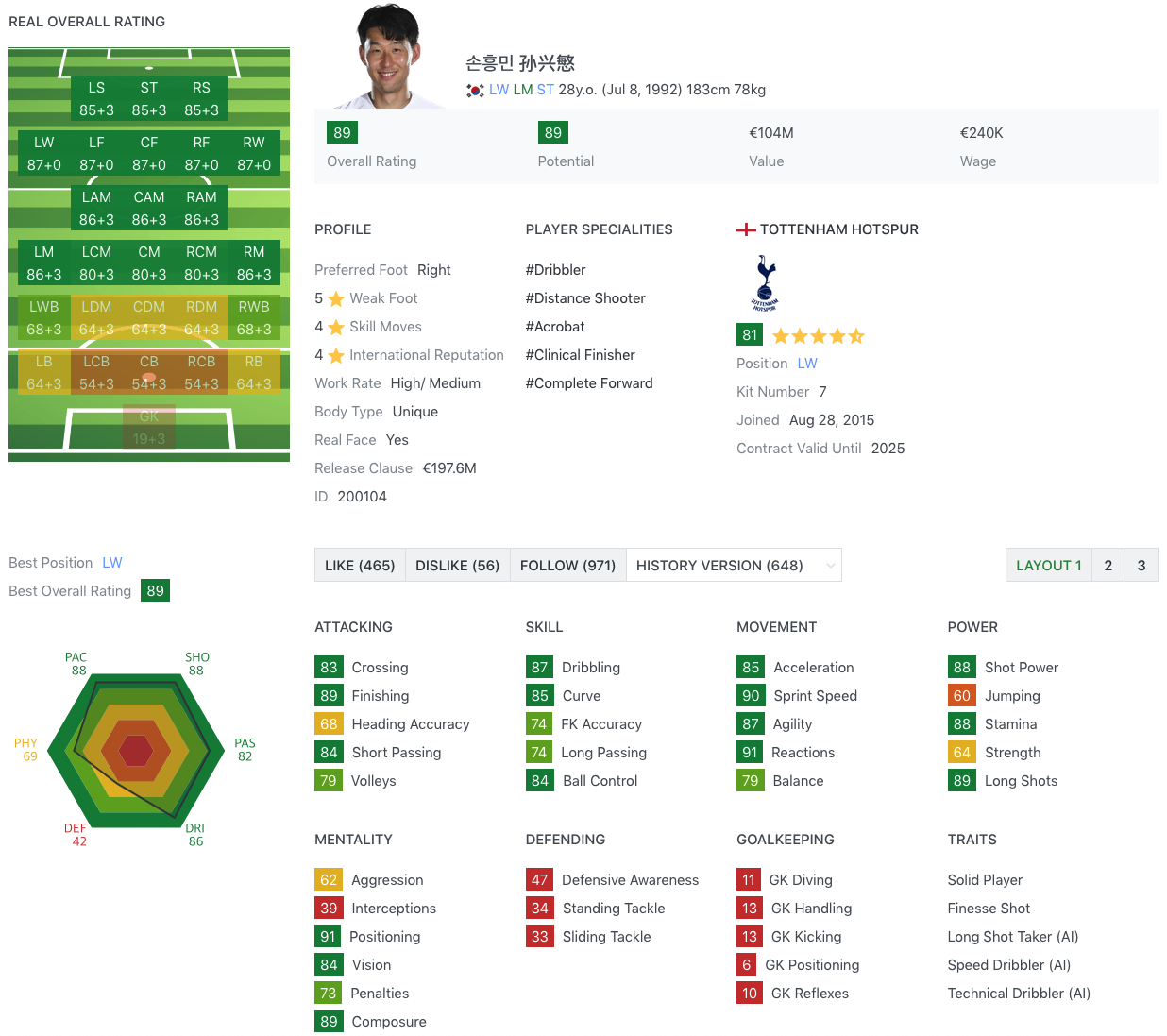}
  \caption{Player (e.g., Son Heung-min) attributes in SOFIFA dataset~\cite{SOFIFA}}
  \label{ability}
\end{figure*}

The SOFIFA dataset includes ability, profile, and position as an attribute type in the game that is quantitatively measured based on a player's actual performance. As shown in Figure~\ref{ability}, ability attributes are interval data showing the player’s soccer performance-related stats from 1 to 99, with a total of 35 ability attributes. In addition, the SOFIFA dataset provides 6 calculated ability attributes on average which are “shooting (SHO)”, “pace (PAC)”, “passing (PAS)”, “dribble (DRI)”, “defending (DEF)”, and “physical (PHY)” by classifying 17 ability attributes out of a total of 35 ability attributes into two or three attributes, and provides the ‘base stats’ attribute by summing the six calculated ability attributes as depicted in Table~\ref{abilityattributes}.  Furthermore, this study classified 35 ability attributes into seven types of calculated ability attributes and extracted the combined values as “attacking”, “skill”, “movement”, “power”, “mental”, “defending”, and “goalkeeping” attributes, and extracted the “total state” attribute by summing the total of 35 ability attributes as shown in Table~\ref{abilityattributes}.

The SOFIFA data provides profile attributes that are real-world data of football players as shown in Figure~\ref{ability}. Profile attributes consist of two types of categorical data (ordinary and nominal data), as shown in Table~\ref{otherattributes}. In profile attributes, “internal reputation (IR)”, “weak foot,” “skill moves,” and “attack/defense work rate,” are ordinal data, and “preferred foot” is the nominal data. Since “preferred foot” is a nominal data type, we used one-hot encoding to dummy values to extract the features. Furthermore, Table~\ref{otherattributes} shows the description of attributes and the range of possible values for each profile attribute.

Position attributes are two attributes (best position and position), which are provided as suitable positions, referring to the history of the position of the actual player in the game among 27 football positions as shown in Figure~\ref{ability}. SOFIFA dataset provides the same position which are "Left Midfielder (LM)", “Left Winger (LW)”, and “Center Forward (CF)” as Son Heung-min plays in the actual game with suitable positions attribute as depicted in Figure~\ref{ability}. SOFIFA dataset is generally provided from at least one to three in consideration of the history of positions played in actual matches by players, mainly with one position for the goalkeeper, and up to three suitable positions for striker, defender, and midfielders. In addition, the position attributes have the “best position” attribute, which is expressed as one of the positions in which the player actually plays the most and does the best in the game of the year. In the case of Son Heung-min, the best position is provided as "Best Position Left Midfielder (BPLM)." Similarly, the position and best position attributes were extracted as features by encoding them with dummy values as categorical data. Detailed description of position attributes is described in Table~\ref{positionattributes}.

As shown in Table~\ref{otherattributes}, “overall rating”, “best overall rating” (BOV), “potential, and “growth” attributes are the ability attributes considered by the several profile attributes. OVE and BOV are rated as the sum of the weighted average of availability attributes (range, 1–99) and international reputation (range, 1–3; calculated by replacing 1 point at 3 points, 2 points for 4 points, and 3 points for 5 points). “potential” attribute is the player’s potential for the current season, and is calculated by adding the player’s “age”, “international reputation”, and the player's actual game history (e.g., goal and assist point) of the player to the overall rating score. Therefore, "potential" is always equal to or higher than "Overall Rating" and has an average of 5 points (range, 0–23 values) higher. Finally, "growth" represents the player's growth this season, minus "overall rating" from "potential".

The SOFIFA dataset, along with the WhoScored dataset, provides demographic and physical attributes, carrier attributes, and monetary value attributes of the actual player's state as shown in Figure~\ref{ability}. Among the demographic attributes, nationality is numerous to describe the country of all players in the five major European leagues as a dummy value. Therefore, we grouped the player’s nationality into five continents, Africa, America, Asia, Europe, and Oceania, and is denoted by a dummy value as shown in Table~\ref{otherattributes}. Next, the units of feet and inches of height are converted to cm, and the units of lbs of weight are converted to kg. In addition, the height and weight were calculated to add the BMI level as a feature. Among monetary value, we used market value data as ground truth data for market value prediction. The average market value of 2,720 players is $9,020.28 \pm 11,299$ k€ (range, 11,299 k€ –105,500 k€). We unify the unit of millions (m)€ or thousands (k)€ to k€ for monetary values (e.g., release clause, market value, and wage).

Additionally, we crawled and extracted attributes from WhoScored regarding the actual match record (e.g., goal points of each player and club, and match points of each club) of 2021–2022 season that were not provided by the SOFIFA 2022 dataset. The extracted club's match record attributes include the goal points of player’s club (goal difference, goal against, and goal acquisition), the winning rate of the player’s club (e.g., victory point, win, draw, and lose), and the team standing (ranking of the player's club) in this season as shown in Table~\ref{otherattributes}. The player's match record which is extracted from WhoScored attributes involve the attack performance attributes (score, assist, goal point, shooting, effective shooting, personal ranking, corner kick, penalty kick), violation attributes (foul, yellow card, red card, offside), and number of games played as depicted in Table~\ref{otherattributes}. The attributes extracted from WhoScored were then merged with each player in the SOFIFA 2022 dataset. Through the data processing and feature extraction process above, this paper finally extracted the 72 attributes (52 ability attributes, five demographic attributes, seven profile attributes, four ability and profile attributes, two monetary value attributes, two position attributes, three goal point attributes, four winning rate attributes, one club ranking attribute, eight club match record attributes, and 13 player match record attributes) for the player's market value prediction.

\begin{table*}
\centering
\caption{Description and range of possible value in calculated ability attributes}
\label{abilityattributes}
\resizebox{\textwidth}{!}{%
\begin{tabular}{lll}
\toprule
\multicolumn{1}{c}{\textbf{\begin{tabular}[c]{@{}l@{}}Calculated Ability\\ Attributes\end{tabular}}} &
  \multicolumn{1}{c}{\textbf{Calculation formula}} & \multicolumn{1}{c}{\textbf{\begin{tabular}[c]{@{}l@{}}Range of \\Possible Value\end{tabular}}}
   \\ \hline\hline
PAC            & (Sprint Speed + Acceleration)/2                                                                         & 1--99    \\
SHO            & (Finishing + Long Shots + Shot Power)/3                                                                 & 1--99    \\
PAS            & (Crossing + Short Passing + Long Passing)/3                                                             & 1--99    \\
DRI            & (Ball Control + Agility + Balance)/3                                                                    & 1--99    \\
DEF            & (Marking + Tackling + Strength)/3                                                                       & 1--99    \\
PHY            & (Strength + Stamina + Jumping)/3                                                                        & 1--99    \\
Attacking &
  \begin{tabular}[c]{@{}l@{}}Crossing + Finishing + Heading Accuracy + Short Passing + Volleys\end{tabular} &
  5--495 \\
Skill          & \begin{tabular}[c]{@{}l@{}}Dribbling + Curve + FK Accuracy + Long Passing + Ball Control\end{tabular} & 5--495   \\
Movement       & \begin{tabular}[c]{@{}l@{}}Acceleartion + Agility + Sprint Speed + Reactions+ Balance\end{tabular}     & 5--495   \\
Power          & \begin{tabular}[c]{@{}l@{}}Shot Power + Jumping + Stamina + Strength + Long Shots\end{tabular}        & 5--495   \\
Defending      & Marking + Sliding Tackle + Standing Tackle                                                              & 3--297   \\
Mentality &
  \begin{tabular}[c]{@{}l@{}}Aggression + Reactions + Positioning +  Interceptions + Vision + Composure\end{tabular} &
  6--594 \\
Goalkeeping &
  \begin{tabular}[c]{@{}l@{}}GK Positioning + GK Diving + GK Handling + GK Kicking + GK Reflexes\end{tabular} &
  5--495 \\
Overall Rating & Overall rating in position                                                                              & 1--99    \\
BOV            & Overall rating in best position                                                                         & 1--99    \\
Base stats     & PAC+SHO+PAS+DRI+DEF+PHY                                                                                 & 6--594   \\
Total stats    & Sum of total 35 ability elements                                                                        & 39--3500 \\ \bottomrule
\end{tabular}%
}
\end{table*}



\begin{table*}
\caption{Description of monetary value, demographic, profile, and match record attributes in 2021--2022 season}
\label{otherattributes}
\resizebox{\textwidth}{!}{%
\begin{tabular}{lllll}
\toprule
\multicolumn{1}{c}{\textbf{\begin{tabular}[c]{@{}l@{}}Types of\\ Attributes\end{tabular}}} & \multicolumn{1}{c}{\textbf{Attributes}} & \multicolumn{1}{c}{\textbf{Description of Attributes}}                                                                                                                    & \multicolumn{1}{c}{\textbf{\begin{tabular}[c]{@{}l@{}}Data\\ Type\end{tabular}}} & \multicolumn{1}{c}{\textbf{\begin{tabular}[c]{@{}l@{}}Range of \\Possible Value\end{tabular}}} \\  \midrule\midrule
\multirow{3}{*}{Monetary value}  
& Market value                            & Football market value of player                                                                                                                                           & ratio                                  & no limitation                                     \\
                                                 & Wage                                    & The weekly salary of a player from affiliated club                                                                                                                        & ratio                                  & no limitation                                     \\
                                                 & Release clauses                         & Buyout clause of player to transfer from affiliated club to another club                                                                                                  & ratio                                  & no limitation                                     \\
\midrule
\multirow{5}{*}{Demographic}                     & Age                                     & Player's age                                                                                                                                                              & ratio                                  & no limitation                                     \\
                                                 & Height                                  & Player's height                                                                                                                                                           & ratio                                  & no limitation                                     \\
                                                 & Weight                                  & Player's weight                                                                                                                                                           & ratio                                  & no limitation                                     \\
                                                 & BMI                                     & Player's body mass                                                                                                                                                        & ratio                                  & no limitation                                     \\
                                                 & Nationally continental                  & \begin{tabular}[c]{@{}l@{}}Continent to which the nationality of the player belongs \\ (Africa, America, Asia, Europe, Oceania)\end{tabular}                              & nominal                                & 0 or 1                                            \\
\midrule
\multirow{11}{*}{Profile}                                                                       & International reputation                & The affiliated club and individual international reputation                                                                                                               & ordinal                                & 1, 2, 3, 4, or 5                                  \\
                                                 & Preferred foot                          & Preferred foot type (Right, Left)                                                                                                                                         & nominal                                & 0 or 1                                            \\
                                                 & five big league                            & \begin{tabular}[c]{@{}l@{}}The type of five big leagues to which the player's team belongs \\ (Spain Primera Liga, Italy Serie A, France Ligue 1, English Premier League)\end{tabular} & nominal                                & 0 or 1                                            \\
                                                 & Weak foot                               & shot power and ball control attributes for other foot than preferred foot                                                                                                 & ordinal                                & 1, 2, 3, 4, or 5                                  \\
                                                 & Skill moves                             & Number of special skills available                                                                                                                                        & ordinal                                & 1, 2, 3, 4, or 5                                  \\
                                                 & Attacking work rate                     & \begin{tabular}[c]{@{}l@{}}The rate of a player's behavior on the pitch in attacking work \\ (Low = 0, Medium =1, High =2)\end{tabular}                                   & ordinal                                & 0, 1, or 2                                        \\
                                                 & Defesive work rate                      & \begin{tabular}[c]{@{}l@{}}The rate of a player's behavior on the pitch in defensive work \\ (Low = 0, Medium =1, High =2)\end{tabular}                                   & ordinal                                & 0, 1, or 2  
                                                 \\
\midrule
\multirow{6}{*}{\begin{tabular}[c]{@{}l@{}}Ability attributes \\considered by\\ profile attributes\end{tabular}}                                                   
                                                 & Overall rating                          & \begin{tabular}[c]{@{}l@{}}Weighted average of ability attributes +  international reputation\\ depending on position\end{tabular}                                        & interval                               & 1--99                                             \\
                                                 & Best Overall rating                     & \begin{tabular}[c]{@{}l@{}}Weighted average of ability attributes +  international reputation\\ depending on best position\end{tabular}                                   & interval                               & 1--99                                             \\
                                                 & Potential                               & \begin{tabular}[c]{@{}l@{}}Player's potential of the current season \\ (overall rating + value considering age, international reputation)\end{tabular}                    & interval                               & 1--99                                             \\
                                                 & Growth                                  & Player's growth of the current season (Potential-Overall Rating)                                                                                                          & interval                               & 0--98                                 
                                                 \\
\midrule 
\multirow{8}{*}{Club's Match Record}            & Goal acquisition               &   Total number of goal scored by our team in season                                                                                                             &   ratio                    &   no limitation                               \\
                                                 & Goal against                          &      Total number loss scored by the opposite team in season                                      &  ratio                    &  no limitation                                         \\
                                                 &   Goal difference    &  Goal acquisition-Goal against &  ratio     & no limitation                                            \\
                                                 & Victory point      &   Total victory point in season         &     ratio                            &  no limitation                                 \\
                                                 & Win point      &    The number of wins in the game of the season          &         interval                     & 1--38                                 \\
                                                 & Draw point                     &   The number of draws in the game of the season         &  interval         &  1--38                                      \\
                                                 & Lose point      &  The number of lose in the game of the season                                   &   interval     &   1--38 
\\
& Team standing                      &    Team rankings for each league in the season     &    ordinal                              &   1--20
\\

\midrule
\multirow{12}{*}{Player's Match Record}            & Scoring point               &        A player's scoring record in the season                   &              ratio                  &        no limitation                          \\
                                                 & Assist point                          &   A player's individual assist record in the season         &      ratio      & no limitation                                            \\
                                                 & Goal point     & Scoring point + Assist point &  ratio                               &  no limitation                                           \\
                                                 & Shooting    &    number of shooting has taken in the game for a season        &    ratio        &      no limitation                             \\
                                                 & Effective shooting     &       number of effective shooting has taken in the game for a season    &   ratio                              &    no limitation                             \\
                                                 & Personal score ranking     &    player's scoring ranking in the season   &     ordinal     &   1--20                                     \\
                                                 & Corner kick    &    The number of corner kick a player has taken in the game for a season        &      ratio      &   no limitation
\\
& Penalty kick                      &    The number of penalty kick a player has taken in the game for a season      &    ratio      &   no limitation
\\
& Foul                      &     The number of times the player was fouled in the game for a season                             &      ratio                           &  no limitation
\\
& Yellow card                      &   The number of times the player was warned in the game for a season           &   interval                       &  1--76
\\
& Red card                      &     The number of times the player was sent off in the game for a season         &    interval     &   1--38
\\
& Offside                      &      The number of times the player offsides in the game for a season                 &    ratio                             &   no limitation
\\

\bottomrule
\end{tabular}
}
\end{table*}

\begin{table*}
\centering
\caption{Description of football position types}
\label{positionattributes}
\resizebox{\textwidth}{!}{%
\begin{tabular}{@{}llll@{}}
\toprule
\textbf{Position Category}   & \textbf{Position Sub-category}   & \textbf{Position Abbreviation} & \textbf{Position Description} \\ \midrule\midrule
\multirow{8}{*}{Attacker} & \multirow{3}{*}{Striker}              & ST  & Striker                      \\
                          &                                       & LS  & Left Striker                 \\
                          &                                       & RS  & Right Striker                \\  \cmidrule(l){2-4}
                          & \multirow{3}{*}{Forward}               & LF  & Left Forward                 \\ 
                          &                                       & CF  & Centre Forward               \\
                          &                                       & RF  & Right Forward                \\ \cmidrule(l){2-4}
                          & \multirow{2}{*}{Winger}               & LW  & Left Winger                  \\
                          &                                       & RW  & Right Winger                 \\ \hline
\multirow{11}{*}{Midfielder} & \multirow{2}{*}{Wide Midfielder} & LM                             & Left Midfielder               \\
                          &                                       & RM  & Right Midfielder             \\ \cmidrule(l){2-4}
                          & \multirow{3}{*}{Attacking Midfielder} & LAM & Left Attacking Midfielder    \\
                          &                                       & CAM & Centre Attacking Midfielder  \\
                          &                                       & RAM & Right Attacking Midfielder   \\ \cmidrule(l){2-4}
                          & \multirow{3}{*}{Central Midfielder}   & LCM & Left Central Midfielder      \\
                          &                                       & CM  & Central Midfielder           \\
                          &                                       & RCM & Right Central Midfielder     \\ \cmidrule(l){2-4}
                          & \multirow{3}{*}{Defensive Midfielder} & LDM & Left Defensive Midfielder    \\
                          &                                       & CDM & Central Defensive Midfielder \\
                          &                                       & RDM & Right Defensive Midfielder   \\ \hline
\multirow{7}{*}{Defender}    & \multirow{3}{*}{Center Back}     & LCB                            & Left Central Back             \\
                          &                                       & CB  & Central Back                 \\
                          &                                       & RCB & Right Central Back           \\ \cmidrule(l){2-4}
                          & \multirow{2}{*}{Full Back}            & LB  & Left Back                    \\
                          &                                       & RB  & Right Back                   \\ \cmidrule(l){2-4}
                          & \multirow{2}{*}{Wing Back}            & LWB & Left Wing Back               \\
                          &                                       & RWB & Right Wing Back              \\ \hline
Goalkeeper                & Goalkeeper                            & GK  & Goalkeeper                   \\ \bottomrule
\end{tabular}%
}
\end{table*}

\subsection{Correlation Analysis}

We identify the features that correlate with the players’ market value. Accordingly, we obtain the correlation values of each feature. Features lists of correlation values 0.4 or more are as follow: 'Release\_Clause', 'Wage', 'Overall', 'Potential', 'Best Composure', 'Short Passing', 'Curve', 'Long Passing', 'Ball Control', 'Vision', ’Total Stats’, 'Base Stats', ’BOV’, ’PAS’, ’DRI’, ’Total Movement’, ’Total Power’, ’Goal Acquisition’, ’Goal Difference’, ’Winning Points’, ’Win’, ’IR’. After that, we compare the difference between the highly correlated feature lists by correlation analysis with ground truth data and the important feature lists (i.e., Indicator of importance for each feature in interpreting ML models) in Section \ref{sec:importance}.


\subsection{Experimental setup}
The experiment was run on the machine with Intel Xeon Gold 6240 2.6GHz CPU, 32GB RAM PC4 2933 MT/s and 64-bit windows 10 operating system with installed Pandas version 1.2.2 and NumPy version 1.20.1.


\subsection{Result and Discussion}
\subsubsection{Evaluation and Validation Metrics}
For the evaluation and validation process, we split the train and test dataset ratio by 20:80 and use only the train dataset in the validation process with $k$-fold cross-validation ($k$ = 10). We utilize mean average error (MAE) and root mean squared error (RMSE) as the evaluation metrics to compare our proposed methods' performances.
%

\begin{figure}
\centering     
\subfigure[GBDT model]{\label{fig:a}\includegraphics[width=0.49\textwidth]{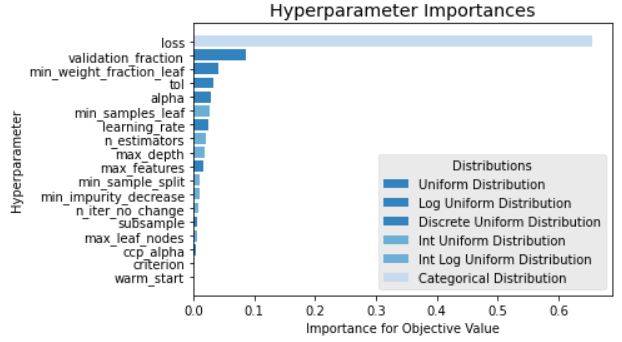}}
\subfigure[LightGBM model]{\label{fig:b}\includegraphics[width=0.49\textwidth]{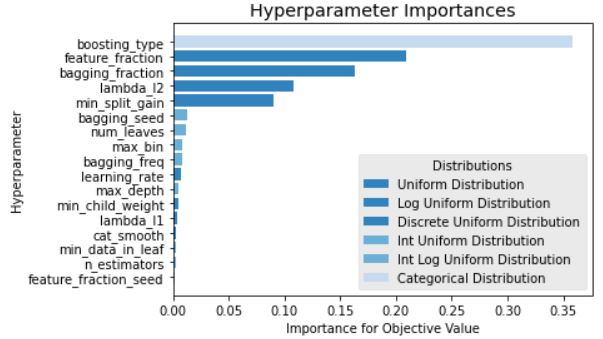}}
\caption{Optimal learning parameters of LightGBM model with highest validation score (RMSE: 716.38) and GBDT model with highest validation score (RMSE: 949.94)}
\label{fig:hyperparameterimportances}
\end{figure}

\begin{table*}
\caption{RMSE and MAE performance comparison of prediction models. Bold text denotes the highest score for each condition: RMSE and MAE for each HPO conditions (Default value, I-TPE, M-TPE)}
\label{results}
\resizebox{1\textwidth}{!}{%
\begin{tabular}{@{}lllllll@{}}
\toprule
& \multicolumn{2}{c}{\textbf{non-HPO (default)}}     & \multicolumn{2}{c}{\textbf{I-TPE}}                 & \multicolumn{2}{c}{\textbf{M-TPE}}                 \\ \cline{2-7}
\textbf{\begin{tabular}[c]{@{}c@{}}\end{tabular}} & \multicolumn{1}{c}{RMSE} & \multicolumn{1}{c}{MAE} & \multicolumn{1}{c}{RMSE} & \multicolumn{1}{c}{MAE} & \multicolumn{1}{c}{RMSE} & \multicolumn{1}{c}{MAE} \\ \hline\hline
\textbf{LM}                                                          & 2,334.65                 & 1,467.45                & \multicolumn{1}{c}{-}    & \multicolumn{1}{c}{-}   & \multicolumn{1}{c}{-}    & \multicolumn{1}{c}{-}   \\
\textbf{Lasso}                                                       & 2,308.03                 & 1,439.23                & 2,232.92                 & 1,349.62                & 2,238.67                 & 1354.32                 \\
\textbf{E-Net}                                                       & 3,222.50                 & 1,859.26                & 2,297.33                 & 1,405.03                & 2,308.56                 & 1,411.48                \\
\textbf{KRR}                                                         & 2,325.09                 & 1,455.31                & 2,321.97                 & 1,451.22                & 2,321.97                 & 1451.22                 \\
\textbf{GBDT}                                                        & \textbf{849.19}          & 417.88                  & 696.73                   & 356.46                  & 1,011.93                 & 546.97                  \\
\textbf{LightGBM}                                                        & 1,069.91                 & \textbf{387.44}         & \textbf{609.42}          & \textbf{211.17}         & 645.55                   & \textbf{239.74}         \\
\textbf{LightGBM (pruning)}                                              & \multicolumn{1}{c}{-}    & \multicolumn{1}{c}{-}   & 636.61                   & 228.93                  & \textbf{632.16}          & 252.57                  \\ 
\bottomrule
\end{tabular}
}
\end{table*}

\begin{figure}
\centering    
\subfigure[]{\includegraphics[width=0.45\textwidth]{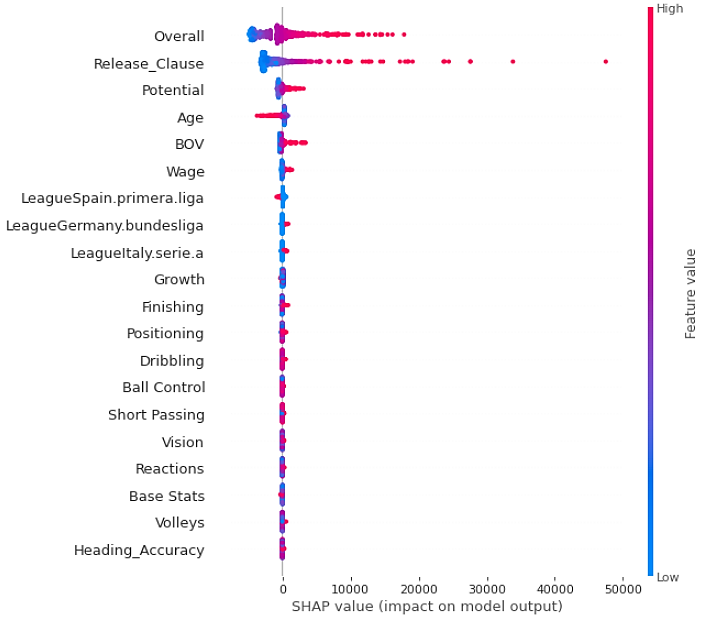}}
\subfigure[]{\includegraphics[width=0.45\textwidth]{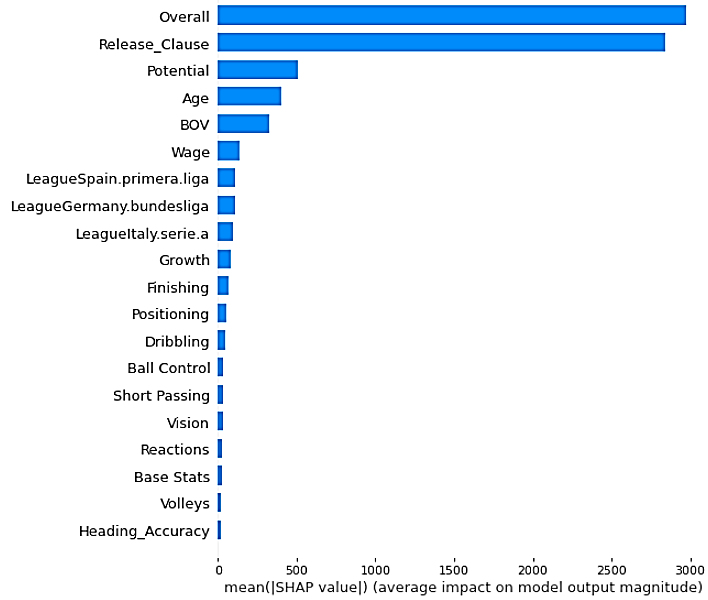}}
\caption{Feature effect (a) and feature importance (b) for best model via SHAP value in ITPE}
\label{fig:featureimportance1}
\end{figure}

\begin{figure}
\centering     
\subfigure[]{\includegraphics[width=0.45\textwidth]{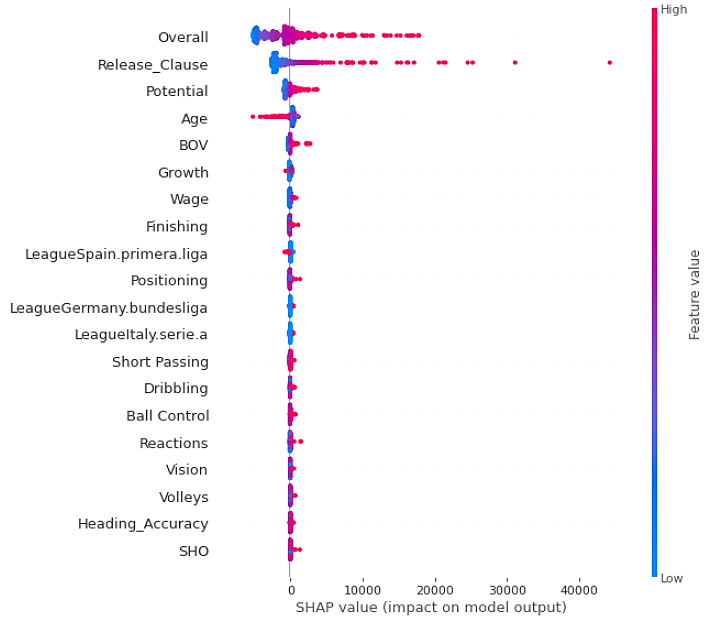}}
\subfigure[]{\includegraphics[width=0.45\textwidth]{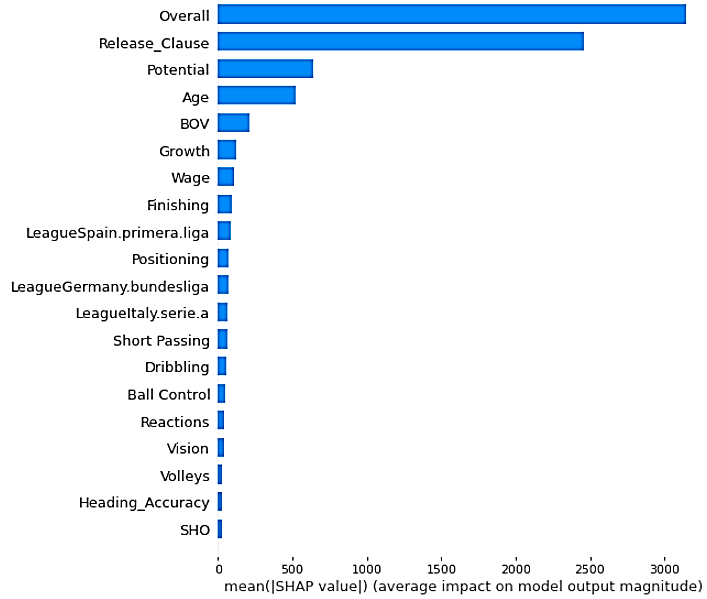}}
\caption{Feature effect (a) and feature importance (b) for best model via SHAP value in MTPE}
\label{fig:featureimportance2}
\end{figure}

\subsubsection{Optimized Hyperparameter Value and Importance}
We obtain 36 HPO results ($3\times6\times2$) with Optuna libraries for six models (i.e., lasso, E-net, KRR, GBDT, LightGBM, and LightGBM with pruning) and two TPE algorithms (e.g., independent TPE, multivariate TPE) to acquire the reliability results of optimized hyperparameter values and feature importance. In addition, we train 100 times for each result. For the regularization regression model, most hyperparameters (e.g., $\alpha$, \emph{tol}, \emph{l1\_ratio}, \emph{coef0}, and $\gamma$) are optimized. For LightGBM and GBDT models, all hyperparameters are searched within recommended search space (e.g., learning rate [search space: 0,1, default: 0.1], n\_estimators [search space: 50, 3000, default: 100] in LightGBM)

To understand which hyperparameters influence the performance, we compare the importance of hyperparameters for the models that representatively show the highest validation performance on each GBDT and LightGBM model. As shown in Figure \ref{fig:hyperparameterimportances}, the \emph{loss} was significantly higher than that of other hyperparameters in GBDT. In case of LightGBM model, the \emph{boosting\_type}, \emph{feature\_fraction}, \emph{bagging\_fraction}, \emph{lambda\_l2}, and \emph{min\_split\_gain} hyperparameters' importance is significantly higher than other hyperparameters. The hyperparameters having high importance in all experiments are \emph{boosting\_type}, \emph{feature\_fraction}, and \emph{bagging\_fraction} in LightGBM model. Morevoer, the loss is relatively higher than other hyperparameters in GBDT model.

\subsubsection{Results of Evaluation}
We divide the results into several cases where HPO is not applied, and HPO is applied using the I-TPE and M-TPE algorithms (see Table \ref{results}). For non-HPO, the performance of GBDT is the highest among all models in terms of the RMSE metric. However, LightGBM shows its superiority regardless of HPO. On average, the performance of LightGBM is 3.8 times and 6.6 times better than the linear and regularization regression model in terms of RMSE and MAE, respectively.

The results indicate that HPO improves the performance of GBDT and LightGBM. However, the linear and regularization regression models (e.g., lasso, E-Net, and KRR) do not benefit from HPO; even in some cases, HPO worsens the performance of such models.
In the I-TPE-based HPO, the LightGBM model shows approximately 1.8 times better performance than the unoptimized GBM model w.r.t both RMSE and MAE metrics. Similarly, in M-TPE-based HPO, unpruned LightGBM outperforms pruned one w.r.t. MAE metric but less significant in terms of RMSE metric. Overall, the I-TPE-based HPO shows more remarkable improvement than the M-TPE-based HPO, demonstrating a difference from previous study~\cite{falkner2018bohb}. Our experimental results also indicate no significant performance difference between unpruned and pruned LightGBM; however, learning cost can be lowered two times when pruning is applied. It supports the previous study, stating that pruning is essential in maintaining the performance of LightGBM and making the learning process efficient.

\subsubsection{Features Importance}
\label{sec:importance}
We identify the importance and effects of the top-20 features out of 124 features that have contributed to the player's value prediction model using the SHAP value (see Figure~\ref{fig:featureimportance1}). In this figure, we plot the best performing model, i.e., the I-TPE-based LightGBM model. As a result, the importance of the features for player's market value is Overall' (i.e., an average of all ability values), 'Release\_Clause' (i.e., price competition system, player's market value of the transfer market, and the amount to the club that owns players can be recruited), 'Age', 'BOV' (i.e., the position with the highest overall ability stats of the player). 

We are interested in comparing the results between correlation analysis and feature importance as shown in Figure~\ref{fig:featureimportance1} and Figure~\ref{fig:featureimportance2}, respectively. Interestingly, 'Release\_Clause' ranks first (e.g., a correlation value of 0.96) in the correlation analysis, but it ranks second in the SHAP-based feature importance. The other two features, such as 'Overall' and 'Potential', also indicate similar patterns. Therefore, it can be concluded that there is a significant correlation between the feature importance result and the correlation analysis result.   


\section{Conclusion}
\label{conc}
This study proposed the application of several state-of-the-art optimized ensemble techniques in sports analytics. This new high-accuracy ML-based predictive model uncovers the potential value of players in the sports field and provides the team with the potential to earn economic returns. Our future study will try to improve the model's performance through another powerful optimized ensemble model (e.g., XGboost, Catboost) via TPE bayesian optimization, which was not used in this study. After that, we will use the stacking ensemble technique (i.e., meta-learning-based ensemble technology that learns how to derive the best performance by well combining multiple models) to combine the extracted optimized GBM, LightGBM, XGboost, and Catboost models. Through this, we will present an advanced ensemble model for prediction with improved efficiency and performance in the sports analytics field.

\bibliographystyle{elsarticle-num-names}
\bibliography{football}

\end{document}